\documentclass[letterpaper, 10 pt, conference]{ieeeconf}  

\IEEEoverridecommandlockouts        

\usepackage{graphics} 
\usepackage{epsfig} 
\usepackage{mathptmx} 
\usepackage{times} 
\usepackage{amsmath} 
\usepackage{amssymb}  
\usepackage[dvipsnames]{xcolor}
\usepackage[space]{cite}    
\usepackage{tabularx}
\usepackage{algorithm}
\usepackage{algpseudocode}
\usepackage{multirow}
\usepackage{caption}
\usepackage[hidelinks]{hyperref}
\usepackage[capitalize]{cleveref}  

\usepackage{setspace}

\definecolor{gentlegreen}{RGB}{20,200,20}
\definecolor{gentlered}{RGB}{200,20,20}

\newcommand{\rev}[1]{\textcolor{black}{#1}}
\newcommand{\finrev}[1]{\textcolor{black}{#1}}

\newcommand{\reg}[1]{{\textsuperscript{\scriptsize \textregistered} #1}}

\title{\LARGE \bf
 Gentle Object Retraction in Dense Clutter Using Multimodal \\ Force Sensing and Imitation Learning
}

\author{\finrev{Dane Brouwer$^{1}$, Joshua Citron$^{2}$, Heather Nolte$^{1}$, Jeannette Bohg$^{2}$, and Mark Cutkosky$^{1}$}
\thanks{\finrev{$^{1}$Department of Mechanical Engineering, Stanford University, USA}}
\thanks{\finrev{$^{2}$Department of Computer Science, Stanford University, USA} \newline
\tt\footnotesize \{\finrev{daneb, jcitron, hanolte, bohg, cutkosky\} @stanford.edu}}%
}

\begin{document}

\maketitle
\thispagestyle{empty}
\pagestyle{empty}

\begin{abstract}
Dense collections of movable objects are common in everyday spaces---from cabinets in a home to shelves in a warehouse. Safely retracting objects from such collections is difficult for robots, yet people do it \rev{frequently, leveraging learned experience in tandem with vision and} non-prehensile tactile sensing on the sides and backs of their hands and arms. We investigate the role of \rev{contact force} sensing for training robots to gently reach into constrained clutter and extract objects.  
The available sensing modalities are (1) ``eye-in-hand'' vision, (2) proprioception, (3) non-prehensile triaxial tactile sensing, (4) contact wrenches estimated from joint torques, and (5) a measure of 
object acquisition obtained by monitoring the vacuum line of a suction cup. 
We use imitation learning to train policies from a set of demonstrations on randomly generated scenes, then conduct an ablation study of wrench and tactile information. We evaluate each policy's performance across 40 unseen environment configurations. Policies employing any force sensing show fewer excessive force failures, an increased overall success rate, and faster completion times. The best performance is achieved using both tactile and wrench information, producing an 80\% improvement above the baseline without force information.
\end{abstract}
\section{Introduction}

Cluttered and constrained environments, like densely packed shelves and cabinets, appear commonly in household and commercial settings. People interact with them daily to place, organize, and extract objects using a combination of tactile sensing and limited vision \cite{vlachou2025tactile,frumento2024unconscious}. 
\rev{Visual information aids in target object identification and provides an initial sense of the obstacle configuration while tactile information can inform when objects are immovable and improve object localization when visual occlusions are present during reaching.}
In such \rev{cluttered} scenarios, there are often no collision-free paths to reach and extract desired objects. Hence, if robots are to operate in these scenarios, they need to embrace contacts, rather than avoid them, as in traditional motion planning \cite{pandey2017mobile}.  At the same time, robots need to be mindful of interaction forces to avoid damage.

\begin{figure}[t!]
\centering
	\includegraphics[width=3.25in]
 {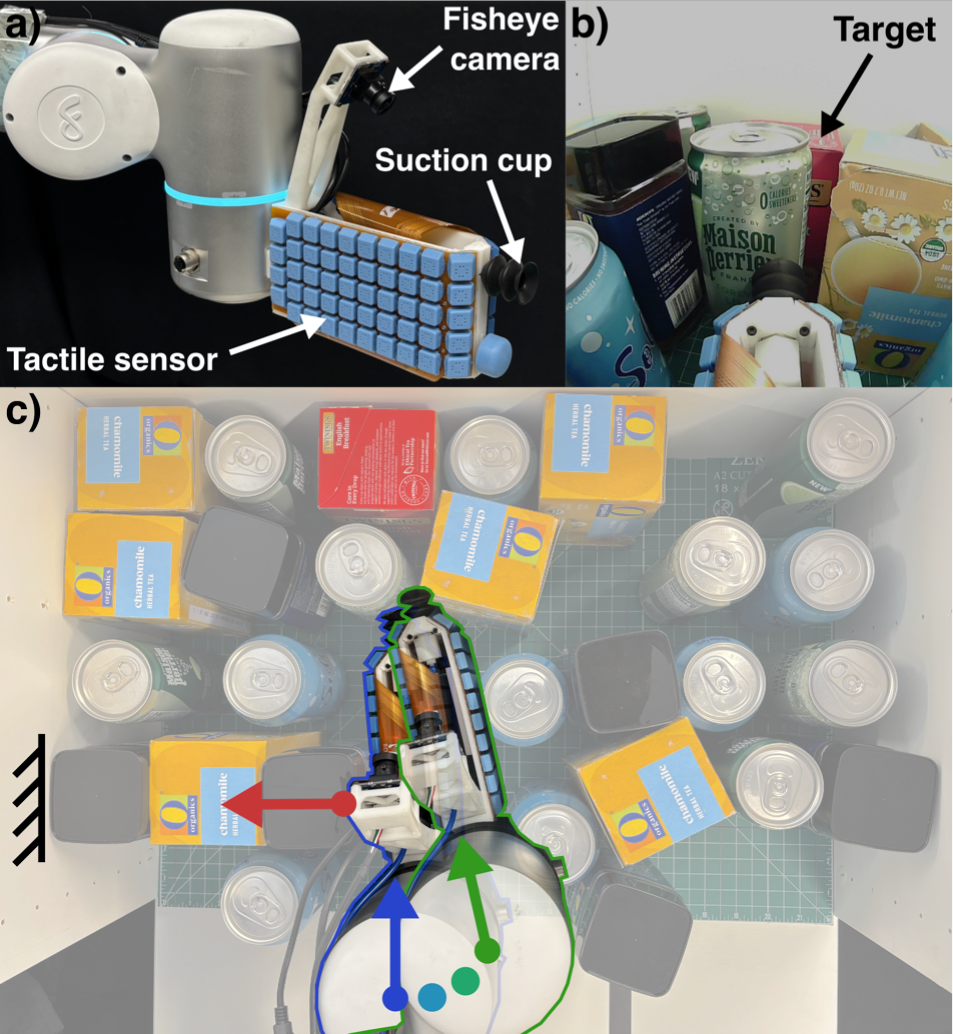}
    \caption{\setstretch{0.9} a) An end-effector equipped with soft, triaxial tactile sensors, a suction cup, and a camera. b) The ``eye-in-hand'' camera view provided as observations to a robot reaching in dense clutter to acquire a target. c) Overhead view (not available to robot) of the scene in (b). Robot's initial pose (blue arrow) causes objects to jam against the left wall, producing a large contact force (red arrow). Action sequence moves the robot to a new pose (green arrow) that is out of contact while still approaching the target.}
    \label{fig:CoverFig}
	\vspace{-15pt}
\end{figure}

Gentle object retraction amidst dense collections of movable objects remains a difficult challenge for most robotic systems, especially in constrained lateral access environments where there is an abundance of visual occlusions, objects cannot be freely pushed aside, and there is no overhead access---as exemplified in \cref{fig:CoverFig}c.
%
%
%
We hypothesize that multimodal vision and force sensing are essential to enable robot policies that embrace contact when it is necessary, but do not cause damage or unecessarily disrupt the environment. 


In this work, we consider the task of gently navigating among dense clutter in a cabinet to reach and extract a target object. 
To complete the task, the robot has access to: (1) ``eye-in-hand'' monocular vision, (2) its own proprioception, (3) contact wrenches estimated from joint torques, (4) signals from soft, triaxial tactile array sensors, and (5) a measure of object acquisition obtained by monitoring the vacuum line of a suction cup. We opt to use imitation learning to provide an unbiased evaluation of how useful the force sensing modalities are for this task.


\paragraph*{Contributions}

We provide an investigation of the relative importance of 
force-torque and tactile sensing for training imitation learning policies to perform gentle, contact-rich manipulation in constrained clutter. We show that wrench and tactile information  \rev{individually} provide performance improvements to robots while performing these tasks\rev{, but we see additional improvements when using both modalities}. \rev{This work demonstrates} object extraction from densely cluttered lateral access scenes (up to \rev{$55$\%} area occupied). 
\rev{While clutter density is an important factor determining task difficulty, additional factors such as the shape variability, deformability, and optical or spectral variability are not explicitly investigated.}

\section{Related Work}
\label{sec:Related Work}

\rev{Utilizing} force sensing in multimodal systems for contact-rich tasks is an active area of research. Efforts include representation learning of wrench or tactile information \cite{lee2020making,guzey2023dexterity}, using kinesthetic demonstrations\cite{ablett2024multimodal,zhang2025kinedex,chen2025dexforce} or hand-held, robot free demonstrations\cite{liu2024forcemimic,liu2025vitamin} for tasks that are hard to teleoperate, and learning to perform bimanual manipulation with visuotactile sensors\cite{zhao2025polytouch,lin2024learning}. These approaches provide investigations of the utility of force sensing for performing dexterous, contact-rich manipulation tasks, but do not explore specifically the problem of highly cluttered environments with contact on non-prehensile surfaces.

Other works perform decluttering\rev{, object retraction,} or packing tasks on tabletop clutter\rev{\cite{huang2021visual,yuan2019end,muhayyuddin2017randomized,srivastava2014combined,kurenkov2020visuomotor, agboh2018real}}, with some approaches using force sensing as part of their multimodal systems\cite{li2022see, murali2022active}. However, in these works the robot can view the manipulation scenario from the top (similar to the overhead view in \cref{fig:CoverFig}c), which does not reflect the visual occlusion encountered in lateral access scenarios. Furthermore, an open tabletop environment avoids the sporadic object jamming that frequently occurs in lateral access scenes where the objects are in an enclosed space. 

Several recent efforts enable object retraction from cluttered lateral access scenes by \rev{reasoning about occlusions and performing mechanical search to isolate target objects \cite{gupta2013interactive, huang2021mechanical, huang2022mechanical, liu2025fetchbot, bejjani2021occlusion}}, yet these approaches are limited to relatively low clutter density, in part due to the lack of force sensing. Specifically, \rev{despite the majority of manipulation maneuvers being non-prehensile, they do not use non-prehensile tactile sensors, which we define as distributed contact force sensors that detect non-prehensile manipulation interactions.}

Non-prehensile tactile sensors have been shown to enable object motion category classification in constrained clutter\cite{thomasson2022going}, but it is non-trivial to incorporate these motion categories into a motion planning system. Prior work has also enabled blind reaching into cluttered environments by leveraging non-prehensile tactile sensing to reach toward target locations amidst fixed obstacles while minimizing contact forces\cite{albini2021exploiting,bhattacharjee2014robotic,jain2013reaching,bohg2010strategies}. This approach, however, limits scene variability and avoids the issue of sporadic changes to object 
impedance when objects jam in constrained yet movable clutter. One recent work uses non-prehensile tactile information to reach toward target locations in constrained, lateral access scenarios by employing motion primitives to react to contacts\cite{brouwer2024tactile}. All of these approaches assume that target locations are known \emph{a priori} and do not support target object extraction. Blind object recognition and retraction have been demonstrated using \rev{both non-prehensile tactile sensing\cite{zhong2022soft} and prehensile tactile sensing\cite{pai2023tactofind}, but these works do} not investigate the interplay of vision and force sensing.

\section{Methods}

To develop an autonomous system that can retract objects from constrained clutter, we develop a custom sensorized end-effector and build on the diffusion policy framework presented in \cite{chi2023diffusion}, leveraging multiple input sensor modalities.

\subsection{Hardware}
\label{sec:hardware}
\begin{figure}[b]
\centering
	\includegraphics[width=3.0in]{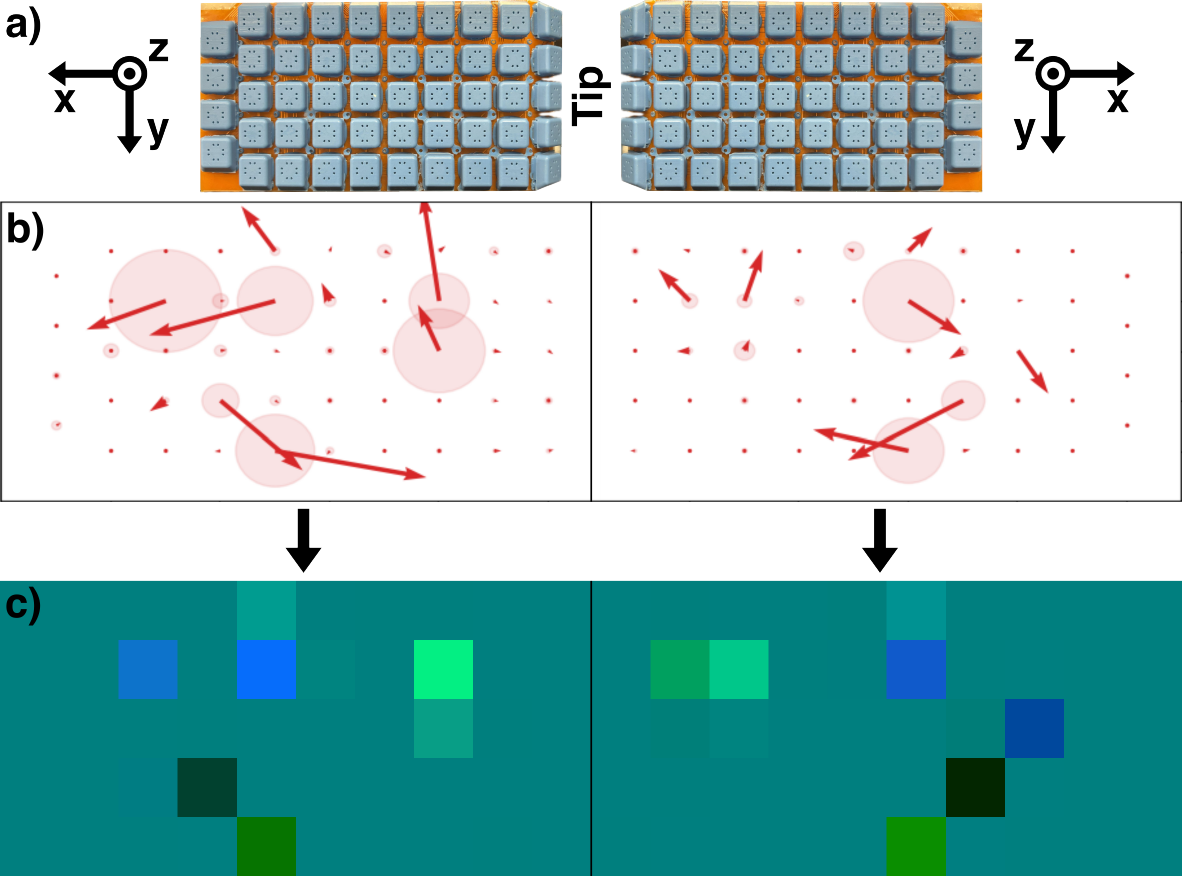}
	\caption{a) Images of the left and right side of the end-effector, along with tactile array coordinate frames, indicating how tactile visualizations map to locations on the end-effector. b) Distributed triaxial force information represented with normal force proportional to circle diameter and shear forces proportional to arrow magnitude and direction.  c) Corresponding tactile image with x, y, and z forces mapped to B, G, and R channels, respectively.}
	\label{fig:TactileRep}
\end{figure}

We use a Flexiv Rizon 4 arm with a 3D-printed end-effector for manipulation within the cabinet. The arm is controlled via a Robot Operating System (ROS) architecture that coordinates between the Flexiv software and each of the sensing modalities. The slender end-effector, as seen in \cref{fig:CoverFig}a, is equipped with a suction cup, an ``eye-in-hand'' fisheye camera that provides images as seen in \cref{fig:CoverFig}b, and two 49-element triaxial tactile sensors as presented in \cite{choi2022deep}. The minimum signal change that the tactile sensors can distinguish from noise while navigating the environment is $\approx0.5\,N$, and the measured triaxial contact forces can be represented with circles and arrows proportional to normal and shear force, respectively, as seen in \cref{fig:TactileRep}b, or as an RGB image, as shown in \cref{fig:TactileRep}c.

\begin{figure*}[t]
\centering
	\vspace{5pt}
	\includegraphics[width=0.9\textwidth]{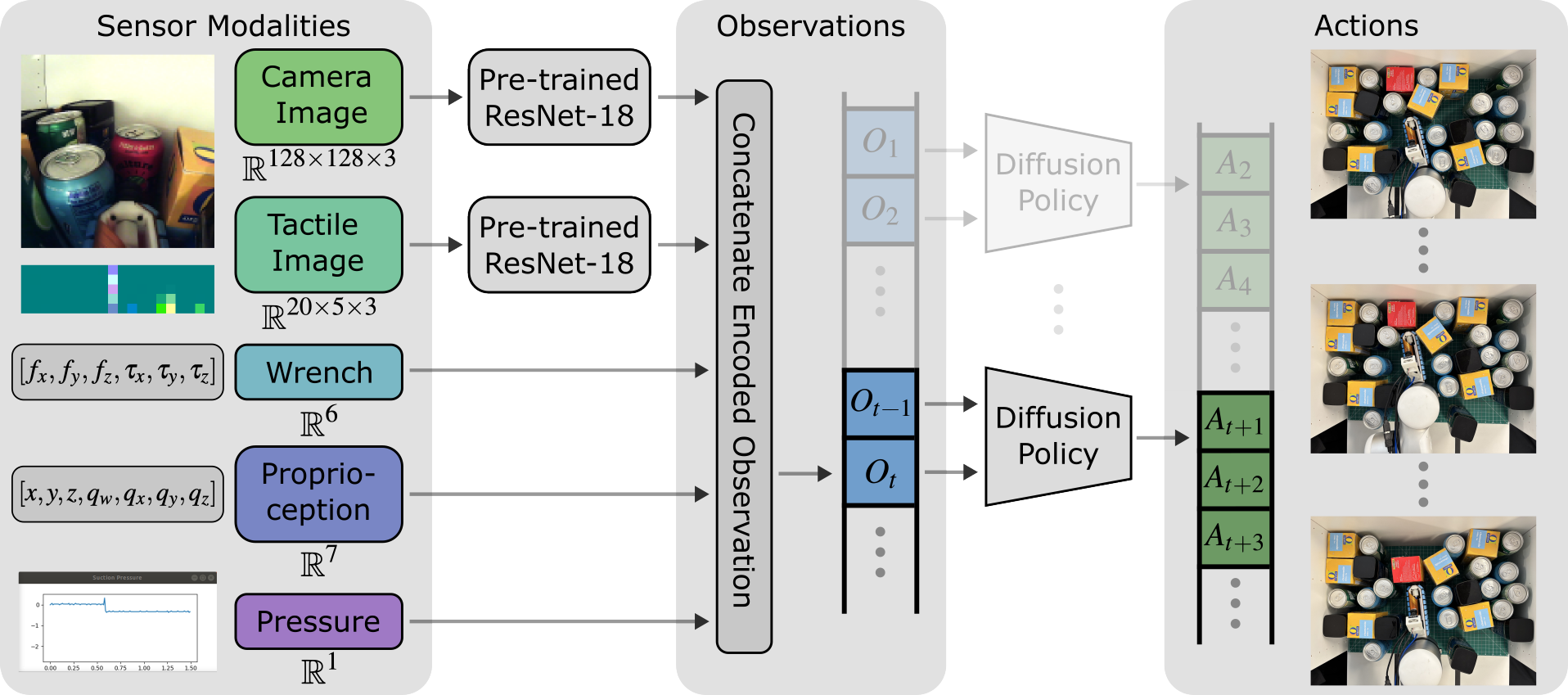}
	\caption{Our force-informed diffusion policy network processes both input images---``eye-in-hand'' camera and tactile---through its own pre-trained ResNet-18 encoder. The output features are directly concatenated with normalized low-dimensional sensor modalities: wrench, proprioception, and pressure. This forms one observation, which is combined with previous observations and passed into a diffusion head to perform action prediction.
    }
	\label{fig:Pipeline}
\end{figure*}

In addition to camera and tactile information, we provide proprioception readings using the robot's joint encoders to form the tool coordinate pose \rev{(TCP), which is located at the suction cup and consists of the cartesian position concatenated with the quaternion orientation}. We also use built-in Flexiv commands that leverage joint-torque sensing to determine the dynamics-compensated wrench, represented in the \rev{TCP} frame. Considering the force components of this wrench (moments are small at the suction cup), the minimum signal change that can reliably be detected above noise due to motion of the arm is $\approx3.3\,N$.

Inspired by \cite{huh2021multi}, we also provide a grasp acquisition metric derived from pressure readings on the suction line. Specifically, we represent pressure as a binary grasp acquisition value that goes high if the pressure reading drops below $-6.9$\,kPa ($-0.07$\,bar) gauge pressure.

\subsection{Training Setup}
\label{sec:training}

Using the hardware described in the previous section, we collect raw sensor data from demonstrations and package them into observations for training imitation learning policies using the diffusion policy architecture\cite{chi2023diffusion}. 

For our training pipeline, we choose to use visually embedded tactile information based on previous findings that reasoning directly on raw high-dimensional tactile information yields worse performance than reasoning on visually embedded tactile information \cite{guzey2023dexterity}.

Thus, we represent tactile information as a $20 \times 5$ pixel image, with each pixel representing one taxel on the end-effector and the three axes of force mapped to the three color channels of the image. First, each tactile array is padded with a single zero-force taxel to make an even 50 taxels. Then, the x forces are mapped from a range of $[-1\,N, 1\,N]$ to pixel intensities $[0, 255]$ on the blue channel. Similarly, the y forces on the left are mapped from a range of $[-1\,N, 1\,N]$ to $[0, 255]$ on the green channel. The y force range for the right sensor is flipped so that upward on the end-effector corresponds to increasing values in both cases. As a result, $0$\,N of shear force corresponds to an intensity of $127$ on both the blue and green channels. The normal z forces are then mapped from $[0\,N, -5\,N]$ to $[0, 255]$ on the red channel. \cref{fig:TactileRep} shows an example of how triaxial force information gets mapped to tactile images. 



\cref{fig:Pipeline} shows the training and inference pipeline used in our system. The $128 \times 128$ pixel ``eye-in-hand'' camera images are fed through a designated ResNet-18 vision encoder, while the $20 \times 5$ tactile images are fed through a separate ResNet-18 encoder. \rev{We choose ResNet since it} was found to be the most effective pre-trained visual encoder for use on tactile information in \cite{guzey2023dexterity} \rev{and its effectiveness was validated in \cite{gu2025tactilealoha}}. After normalization, pose, pressure, and 
wrench are concatenated and fed as separate, low-dimensional inputs to the diffusion policy framework. 
Diffusion policy then outputs actions that can be executed by the robot controller. \rev{The action space at timestep $i$ is \finrev{an 8-dimensional vector} of the form}
\begin{equation} \color{black}
    A_i = [x_i, y_i, z_i, q_{x,i}, q_{y,i}, q_{z,i}, q_{w,i}, g_i]
\end{equation}
\rev{where $[x_i, y_i, z_i]$ is the commanded cartesian position, $[q_{x,i}, q_{y,i}, q_{z,i}, q_{w,i}]$ is the commanded quaternion orientation, and $g_i$ is the binary gripper command that controls suction.}

\section{Data Collection}
\label{sec:Data}
To build a dataset of gentle object retraction in clutter, we design a system to generate randomized cluttered environments, outline success and failure metrics for our task, and use a teleoperation setup for gathering expert demonstrations. 

\subsection{Environment Setup}
\label{sec:env}

\begin{figure}[!t]
\centering
	\vspace{5pt}
	\includegraphics[width=3.25in]{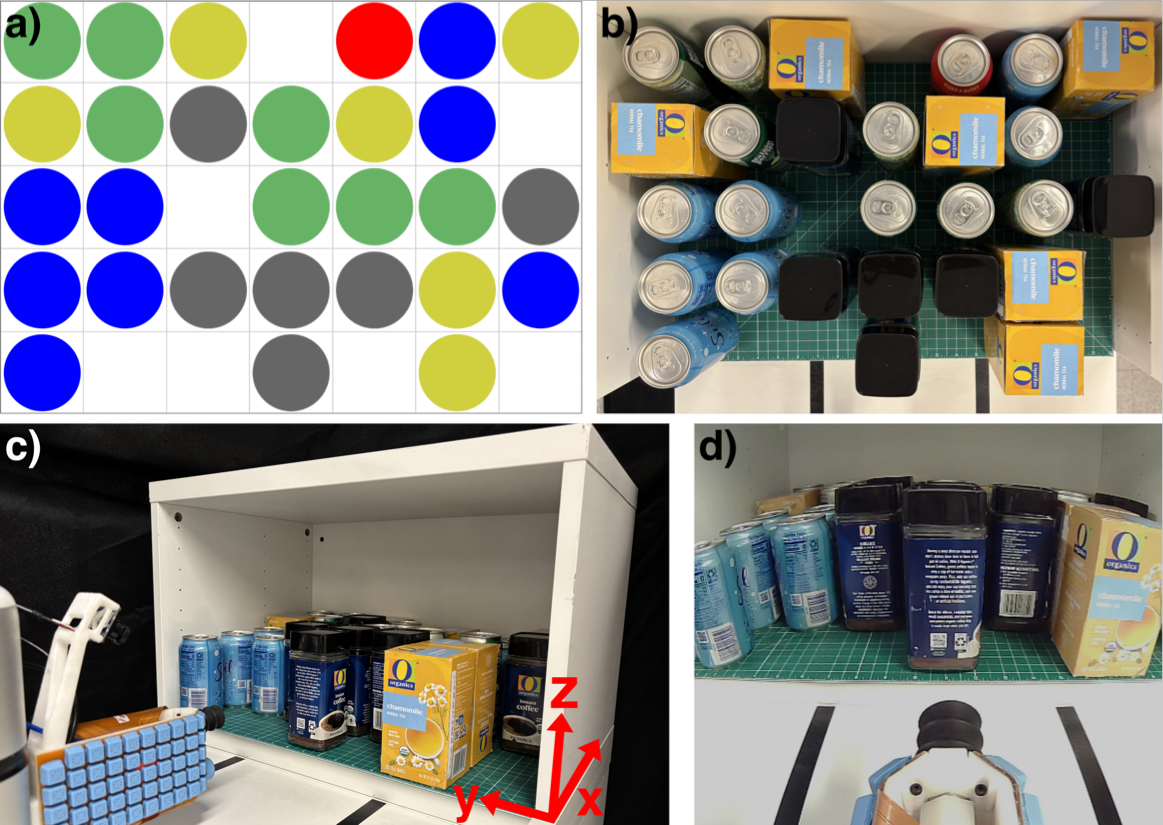}
	\caption{a) Schematic of randomly generated configuration of cluttered objects with corresponding b) overhead view of the physical configuration, c) external view of the scene, and d) ``eye-in-hand'' camera view provided to the robot. Each scene has 25-28 obstacles of 4 types---blue, green, black, and yellow---and one of 3 possible red target objects.}
 
	\label{fig:SceneSetup}
\end{figure}

We construct a cluttered environment using a 5$\times$7 grid of possible object locations within a 38\,cm $\times$ 53\,cm cabinet shelf that has a height of 32\,cm. \cref{fig:SceneSetup}a shows an example of a generated schematic that is then used to populate the physical environment, as seen in \cref{fig:SceneSetup}b. The same scene is also shown from an external view in \cref{fig:SceneSetup}c and from the robot's ``eye-in-hand'' view in \cref{fig:SceneSetup}d. We first select one of three possible red target objects---a cardboard tea box \rev{(area of $55.87\,cm^2$)} and two shapes of aluminum cans \rev{(areas of $34.25\,cm^2$ and $25.87\,cm^2$)}---and place it in one of the middle three grid cells at the back of the cabinet. \rev{Scenes which contain a target object in the back corner may need to consider alternative strategies, such as multiple suction attempts to move impeding obstacles aside.} Each remaining grid cell can be empty or filled with one of four object types: blue, green, black, or yellow, which have footprint areas of 34.26\,$\text{cm}^2$, 26.32\,$\text{cm}^2$, 40.97\,$\text{cm}^2$, and 57.03\,$\text{cm}^2$, respectively. Each randomly-generated scene then populates between 25 and 28 obstacles, using a uniform random sampler for all selections. This procedure results in the least cluttered scenes having \rev{$45\%$} area occupied and the most cluttered scenes having \rev{$55\%$} area occupied. The cluttered environments generated in this way are highly variable, with $\approx10^{20}$ possible discrete object configurations.

\subsection{Task Setup}
\label{sec:task}

The task involves starting at a home position in front of the cabinet shelf with a goal of gently navigating through the cluttered environment to acquire a target object and bring it out of the shelf. Successful completion of the task requires locating the correct target object (the only red object in the scene), retracting it to the home position, and dropping it into the region outlined in black (visible beneath the suction cup in the foreground of \cref{fig:SceneSetup}c,d). We require that the task is completed within 120 seconds. \rev{Since we use a sampling rate of $f_s = 10\,Hz$, timeout failures will result in an episode length that contains $\approx1,200$ action steps, whereas successful trials will fall into a distribution of shorter episodes.}

This task fundamentally requires embracing contacts as there is almost never a contact-free path to the target object, yet the constrained nature of the environment means that objects are likely to impede robot motion, which can cause damage to the environment or the robot if the non-prehensile pushing maneuvers are not performed carefully. To ensure safe behaviors, we require that interactions limit the momentum transfer to the environment. In other words, motions cannot surpass an excessive net force impulse, $I_{net}$

\begin{equation}
    I_{net} = F_{net} \delta t_{react}
\end{equation}
where \rev{$\delta t_{react}$ is the time given to both demonstrator and policy to react to the contact and $F_{net}$ is the L-2 norm of the first 3 elements of the measured wrench averaged over the previous duration of $\delta t_{react}$.} We also require that motions cannot exceed a peak force impulse, $I_{peak}$

\begin{equation}
    I_{peak} = F_{peak} \delta t_{react}
\end{equation}
where $F_{peak}$ is the maximum L-2 norm of the triaxial tactile forces across all taxels. \rev{This value of $F_{peak}$ is also averaged over the previous duration of $\delta t_{react}$ at each timestep.}

We implement two distinct impulse thresholds because the 
peak
force is the metric most likely to damage an object and the wrench measurements are unaware of individual contact magnitudes when there are multiple simultaneous contacts. The peak tactile force, however, does not have full coverage of the robot arm and will be unaware of some contacts, thus a net force threshold is also needed.

We determine $F_{net}$ and $F_{peak}$ values for our task by teleoperating the robot into 2 fragile objects---a plastic cup and the yellow cardboard tea box used in our scenes---until substantial damage occurs. In the case of the tea box, operation was stopped once the cardboard buckled or tore and in the case of the plastic cup, operation was stopped once the cup cracked. We repeat this process 3x for each, then extract the largest value of net force and peak tactile force. We then average these values across all trials and round up to the nearest integer, resulting in a net force magnitude threshold, $F_{net}$, of $26$\,N and a peak tactile magnitude threshold, $F_{peak}$, of $6$\,N. While we recognize that these forces are not applicable to all possible tasks, they serve as reasonable thresholds to determine gentleness as it pertains to our task and environment.

We select $\delta t_{react}$ according to the standard number of action steps, $n_{act} = 8$, in the diffusion policy framework, in combination with the sampling rate, $f_s = 10\,Hz$, used in our system. In this way, $\delta t_{react} = 0.8\,s$ enables the autonomous policies to properly react to contacts in time. Qualitatively, the demonstrator also needed this reaction time in order to properly avoid exceeding the impulse thresholds during teleoperation. 

Through this process, we establish the excessive impulse task constraints of $I_{net} < 20.8\,Ns$ and $I_{peak} < 4.8\,Ns$. \rev{Adjusting these thresholds for alternative tasks merely requires repeating these initial safety tests on the new scene.}

\subsection{Demonstration Setup}
\label{sec:demos}

\begin{figure}[t!]
\centering
	\vspace{8pt}
	\includegraphics[width=3.4in]{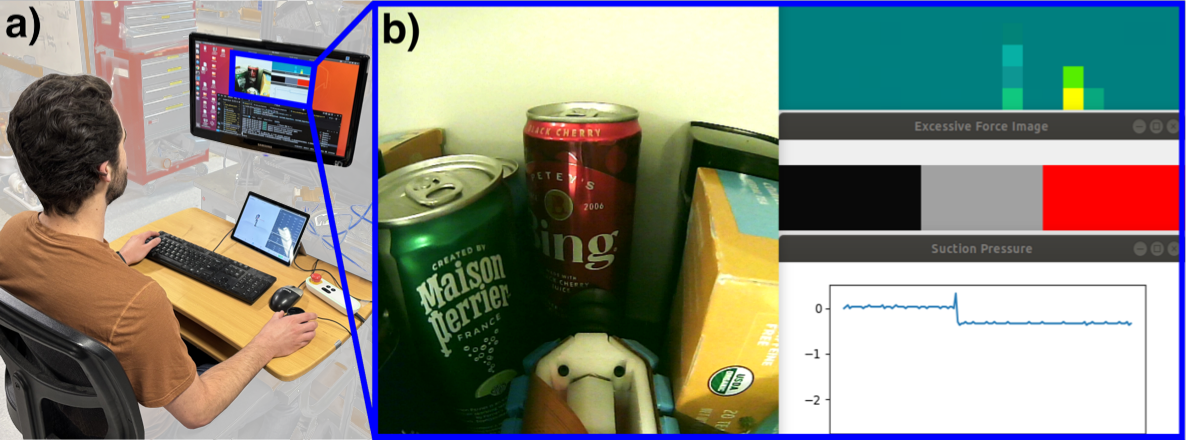}
	\caption{a) Demonstrator using a spacemouse to control robot motion while observing b) visual, tactile, excessive force warnings, and pressure information provided on screen. The contact on the right side of the end-effector, which is unseen in the camera view, is causing a large peak force as indicated by the red rectangle. The demonstrator does not have an external view of the physical scene.}
 
	\label{fig:demos}
\end{figure}

Using a 3Dconnexion SpaceMouse\reg{}Compact to teleoperate the robot, we collect demonstrations on randomized cluttered scenes generated through the process described in \cref{sec:env}. \rev{The teleoperation is performed by mapping the $x$, $y$, $z$, and $yaw$ signals of the SpaceMouse to incrememental changes in the $x$, $y$, $z$, and $\theta_z$ values of the robot's commanded TCP. The $\theta_x$ and $\theta_y$ values are fixed to maintain alignment with the cabinet. The left button of the SpaceMouse turned suction on by switching a 3-way solenoid valve to the vacuum pump, whereas the right button switched the solenoid valve to atmospheric pressure, quickly releasing the object.} As seen in \cref{fig:demos}, the demonstrator has access to the ``eye-in-hand'' camera view, the tactile images as described in \cref{fig:TactileRep}, and pressure sensing, as well as an additional ``excessive force'' image which visually encodes whether the robot is approaching $F_{net}$ and $F_{peak}$.
The left and right sides of this excessive force image (shown in \cref{fig:demos}b) represent peak tactile force from the left and right sides of the end-effector while the center portion corresponds to the net force magnitude computed using the joint torques. In these portions, we display grayscale values ranging from black to white as the corresponding force values range from $0$\% to $80$\% of $F_{net}$ and $F_{peak}$. Beyond $80$\%, we display a red warning in the corresponding portion to enable the demonstrator to respond. This visualization system was empirically tuned to provide the demonstrator with enough warning to avoid excessive force failures while avoiding false positives as necessary non-prehensile maneuvers are performed. 

\rev{To maintain a consistent demonstrated policy that spans a range of strategies, the following criteria were used to determine actions during teleoperated task completion. If the operator saw the target object to start the task, progress directly toward the target was commanded. If the target object was not initially visible, each task would alternate whether initial progress was commanded forward and left (favoring the $+y$ direction) or forward and right (favoring the $-y$ direction). Once the target became visible, the teleoperator would resume progress toward the target object. While making progress directly toward the target object, the operator pushed impeding obstacles aside to clear space, resulting in a roughly sinusoidal, snaking motion. During these pushing maneuvers, if obstacles jammed against the environment constraints, the excessive force indicator warned the teleoperator to move away from the jammed cluster of obstacles and reorient to maintain focus on the target object. If object jamming persisted with no clear path remaining toward the target, the operator retracted to the opening of the lateral access scene and re-entered a new location to break apart the jammed cluster of obstacles. Once the target was isolated, the demonstrator acquired the target object and retracted it to the final goal position.}



\section{Experiment}
\label{sec:experiment}

To investigate the hypothesis that robots need force sensing for gentle object retraction in constrained clutter, we conduct a force ablation experiment by training and evaluating four policies with varying access to wrench and tactile information.

\subsection{Force Ablation Experiment}
Using the setup described in \cref{sec:demos}, we gather 100 demonstrations, gently navigating the scene to acquire the red target object. If either $I_{net}$ or $I_{peak}$ is exceeded, the demonstration automatically ends and the scene has to be re-collected by the demonstrator.

With these demonstrations and the architecture described in \cref{sec:training}, we trained four policies with varied access to wrench and tactile information to investigate their importance for this task: 1) the \textit{baseline} policy masks out (replaces with zero values) wrench and tactile information, 2) the \textit{wrench-informed} policy masks out tactile information, 3) the \textit{tactile-informed} policy masks out wrench information, and 4) the \rev{\textit{wrench\,+\,tactile}} policy has both wrench and tactile information included. Apart from this masking, all policies are trained using the same architecture and hyper-parameters, with each policy trained for $200$ epochs.

To ensure that the policies have access to the same information they are trained on, we similarly mask out the wrench or tactile information during evaluations for each policy.
We first generate a randomized scene as described in \cref{sec:env}, then evaluate the performance of each of the 4 policies on the same scene according to the success and failure criteria described in \cref{sec:task}.

\subsection{Results}

\begin{figure}[t]
\centering
	\vspace{8pt}
	\includegraphics[width=3.4in]{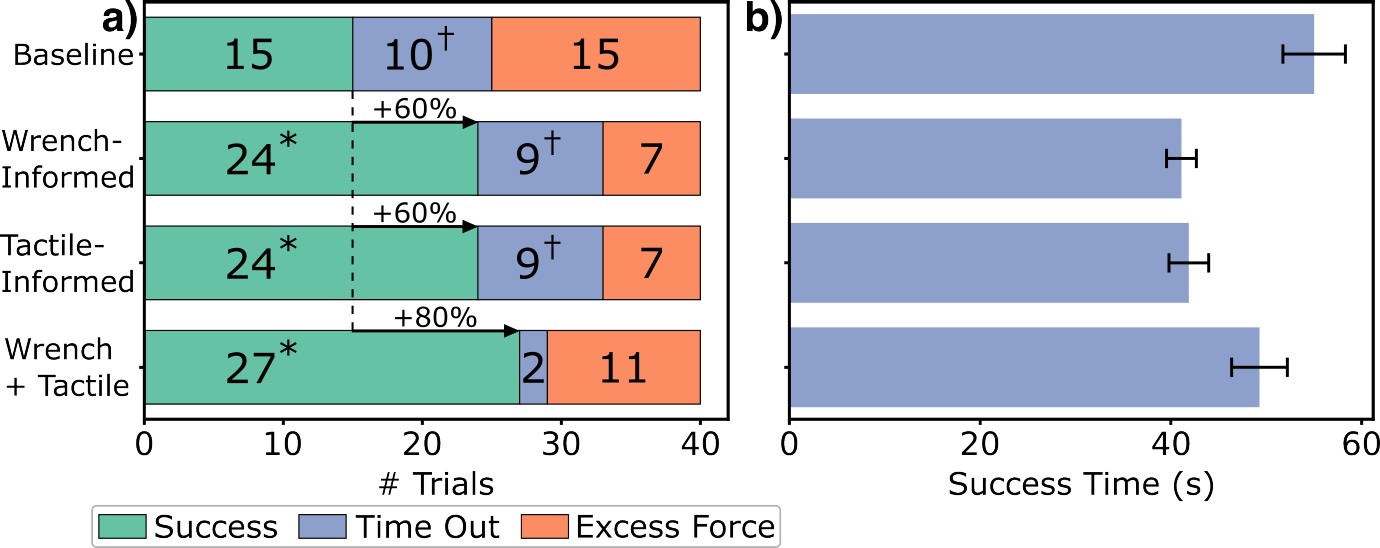}
	\caption{Performance results for 40 evaluations of the force ablation experiment. a) Proportions of successes, time out failures, and excessive force failures for each of the policies. \textsuperscript{*} indicates $p<0.05$ in a two-sided pairwise z-test against the \textit{baseline}. \textsuperscript{\textdagger} indicates $p<0.05$ in a two-sided pairwise z-test against the \rev{\textit{wrench\,+\,tactile}} policy. b) Success times for each policy with error bars showing standard error. Each policy with any force modality improved success rate and accomplished the task faster.}
 
	\label{fig:results}
\end{figure}

The results of $40$ evaluations on each policy are summarized in \cref{fig:results}. The first item of note is that each policy that has access to wrench or tactile information in addition to vision significantly outperforms the \textit{baseline} policy. 
This can be seen in \cref{fig:results}a, which displays the number of successes, time-out failures, and excessive force failures for each policy. The best performance is demonstrated by the \rev{\textit{wrench\,+\,tactile}} policy, where we see an $80$\% relative improvement above the baseline. The improved performance for all policies with access to either force modality
appears to be largely due to a substantial reduction to excessive force failures. 

The \rev{\textit{wrench\,+\,tactile}} policy appears to derive its improved performance largely from the ability to avoid time-out failures as seen in \cref{fig:results}a. In contrast, the significantly higher proportion of timeout failures for the remaining policies often occurred due to pulling away from objects in the absence of large contact forces, failing to focus on the target after it has been isolated, and even retracting without the target object. \rev{These behaviors are shown in the accompanying video in the supplementary materials.} The \textit{baseline} policy exhibits the highest timeout failure rate. Finally, as seen in \cref{fig:results}b, all policies informed by wrench or tactile information accomplished the task faster, with the \textit{wrench-informed} policy being fastest. 


\begin{figure}[!t]
\centering
	\vspace{5pt}
	\includegraphics[width=3.25in]{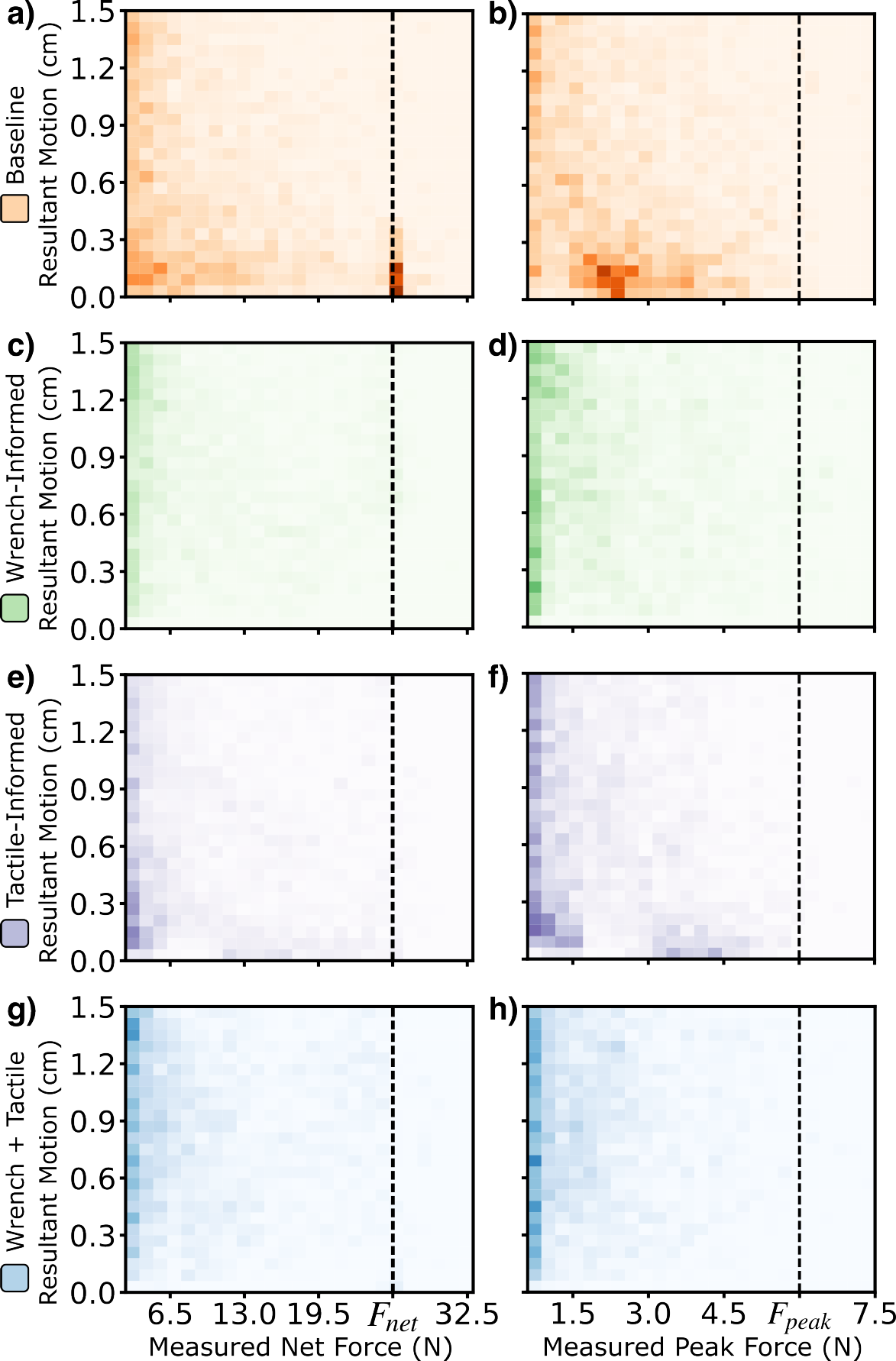}
	\caption{Heatmaps of resulting motion after \rev{measured} net and peak forces for \rev{all} policies. a) Motion performed by the \textit{baseline} policy as a function of \rev{measured} net forces. b) Resulting motion versus \rev{measured} peak forces for the \textit{baseline} policy. \rev{c) Resulting motion as a function of measured net force for the \textit{wrench-informed} policy. d) Resulting motion as a function of measured peak force for the \textit{wrench-informed} policy. e) Resulting motion as a function of measured net force for the \textit{tactile-informed} policy. f) Resulting motion as a function of measured peak force for the \textit{tactile-informed} policy. g) Resulting motion as a function of measured net force for the \textit{wrench\,+\,tactile} policy. h) Resulting motion as a function of measured peak force for the \textit{wrench\,+\,tactile} policy. The black dashed lines represent the excessive force values used for our task. The \textit{baseline} policy displays the highest proportion of large force, low motion datapoints, indicating a failure to react to contacts.}}
 
	\label{fig:ForceEnv}
\end{figure}


To gain a better understanding of how the policies have learned to react to contact forces, we extract the \rev{measured} forces and resulting displacements of the end-effector for all timesteps of all evaluations \rev{for each policy}. We compute the resulting displacement over a subsequent duration of $0.8\,s$ to allow for 1 full inference cycle to enable proper reactions. 

The observed motion versus net forces and peak tactile forces can be seen in \cref{fig:ForceEnv}, where cells toward the bottom right of each plot correspond to a large observed force but little resulting motion---indicating a failure to respond. For reference, the values of $F_{net}$ and $F_{peak}$ are plotted as vertical \rev{dotted} lines. Ideally, there should be no cells that correspond to little or no motion when forces exceed or approach these values, since remaining in place would maintain the excessive force and cause an impulse failure. 
As seen in \cref{fig:ForceEnv}a,b, the \textit{baseline} policy 
contains a substantial number of high force, low motion datapoints, but the \rev{remaining policies in \cref{fig:ForceEnv}c-h largely avoid low motion as the observed forces increase. Notably, the \textit{tactile-informed} policy appears to have more datapoints at large peak force and low motion than either the \textit{wrench-informed} or \textit{wrench\,+\,tactile} policies, indicating that wrench sensing appears to be the most effective modality for reacting to large forces.} 

We also note the number of failed reactions by tallying the total instances where less than $0.5\,cm$ of displacement is seen after observing either a net force that exceeds $F_{net}$ or a peak tactile force that exceeds $F_{peak}$. The \textit{baseline} policy fails to react to large contact forces $412$ times in contrast to \rev{$17$ failed reactions for the \textit{wrench-informed} policy, $72$ for the \textit{tactile-informed} policy, and $22$ for the \textit{wrench\,+\,tactile} policy}.

\section{Discussion}

When attempting to accomplish object retraction in constrained clutter using imitation learning, we have demonstrated that the addition of force sensing modalities intuitively improves a policy's ability to be gentle. Beyond this, we note that each policy in our experiment which lacked any force modality that the demonstrator used to react saw a significant increase in timeout failures\rev{, during which the policies displayed erratic behavior}. \rev{This may indicate that the performance improvements seen with the addition of contact force sensing result from enabling policies to correctly identify the cause-effect relationship between observations and expert actions. Previous work establishes a framework to interpret causal relationships in imitation learning by supposing that there are disentangled factors of the observation data, some of which are true causes for expert actions and some of which are nuisance factors that coincidentally correlate with expert actions\cite{de2019causal}. This prior work demonstrates that the 
removal of nuisance factors from the observation data can reduce causal confusion in imitation learning. 
Since highly cluttered and occluded environments have many visual distractors that do not correlate to expert actions, our experiments suggest that the addition of true causes of expert actions---in our case, contact force sensing---may be an alternative method of reducing causal confusion amidst visual distractors, though future experimentation is needed to verify this correlation.}

\rev{It is worth noting that the \textit{wrench\,+\,tactile} policy exhibited a higher excess force failure rate than the \textit{wrench-informed} and \textit{tactile-informed} policies. While this counter-intuitive result is not statistically significant and may be a result of chance, an additional explanation could be that policies containing more observation data modalities are often less data efficient, requiring more training data to achieve satisfactory performance
\cite{hao2023masked, kumar2025collage}. Since the number of demonstrations used in our study is relatively low, the \textit{wrench\,+\,tactile} policy may suffer from excess force failures due to this data efficiency issue.}

As robots increasingly delve into unstructured environments, we should expect that conventional visuo-proprio policies will suffer due to an inability to react to unexpected contacts not only at the grasping surfaces of the robot, but also along non-prehensile surfaces. This work highlights the need of both multi-modality and whole-body sensing, especially if we seek to enable robust manipulation in unstructured human environments.

\section{Conclusion \& Future Work}
In this work, we demonstrate the ability to use imitation learning to train autonomous policies for gently navigating and retracting objects from extremely variable, densely cluttered lateral access environments. 
We find that in contrast to a baseline policy without access to wrench and tactile information, any policy with access to one or both force modalities demonstrates an improvement in overall success rate, a reduction in excess force impulse failures, and an improvement to the average success time. The \rev{\textit{wrench\,+\,tactile}} policy shows the best performance, with an improvement of $80$\% over the baseline.

Future work in this area includes improving demonstration data by providing the demonstrator with a kinesthetic haptic feedback teleoperation device and a haptic sleeve (e.g. as in \cite{du2024haptiknit})
to supply wrench and tactile feedback, respectively. We believe this would enable demonstrators to respond more naturally to contact forces, likely leading to more effective trained policies. The hardware could also be adapted to better capture the potential of non-prehensile tactile sensing by building in more degrees of freedom for reacting to tactile stimuli. The end-effector could be replaced with a multifingered manipulator, though this would require more complex tactile sensing geometries and the development of a new grasp acquisition metric. 

\rev{Furthermore, an additional study of the system's robustness to out of distribution objects for both obstacles and targets could prove useful to further investigate the importance of tactile and wrench information in these scenarios. Using a pre-trained segmentation network may be necessary to identify arbitrary objects rather than only red targets.} \rev{To alleviate the limitation that the force thresholds used in this work are task specific, adaptive gentleness could be implemented, perhaps with control barrier function approaches \cite{lu2024robot,vinter2024safe}.} The network architecture could be improved by fine tuning policies with reinforcement learning as in \cite{haldar2023teach}, incorporating more reactive diffusion policy architectures as in \cite{xue2025reactive}, or building a custom tactile encoder from play data as in \cite{guzey2023dexterity}. The proposed imitation learning framework could also be compared against a system that implements a 
model-based, low-level force-informed controller as in \cite{killpack2016model}
integrated with learned motion planning using vision and proprioception. \rev{Investigating whether the benefits of contact force sensing modalities reported in this work are echoed in alternative control architectures would be an especially valuable future direction.}

More broadly, this work represents a call to move beyond robot interaction in mere visual clutter and to embrace the degree of physical clutter encountered on shelves and in cabinets in homes and commercial settings. By learning from the \finrev{embodied}, sensing, and reactive strategies that humans use on a daily basis, we can enable robots to perform robust, contact-rich manipulation in unstructured environments.

\addtolength{\textheight}{-1cm}   





\section*{ACKNOWLEDGMENT}

\finrev{Toyota Research Institute provided funds to support this work. D. Brouwer was supported by a Stanford Graduate Fellowship. The authors thank Rick Cory, Hongkai Dai, Hojung Choi, Marion Lepert, and Claire Chen for insightful discussions.}



\bibliographystyle{IEEEtran}
\bibliography{References}

\begin{thebibliography}{10}
\providecommand{\url}[1]{#1}
\csname url@rmstyle\endcsname
\providecommand{\newblock}{\relax}
\providecommand{\bibinfo}[2]{#2}
\providecommand\BIBentrySTDinterwordspacing{\spaceskip=0pt\relax}
\providecommand\BIBentryALTinterwordstretchfactor{4}
\providecommand\BIBentryALTinterwordspacing{\spaceskip=\fontdimen2\font plus
\BIBentryALTinterwordstretchfactor\fontdimen3\font minus \fontdimen4\font\relax}
\providecommand\BIBforeignlanguage[2]{{%
\expandafter\ifx\csname l@#1\endcsname\relax
\typeout{** WARNING: IEEEtran.bst: No hyphenation pattern has been}%
\typeout{** loaded for the language `#1'. Using the pattern for}%
\typeout{** the default language instead.}%
\else
\language=\csname l@#1\endcsname
\fi
#2}}

\bibitem{vlachou2025tactile}
M.~E. Vlachou, J.~Legros, C.~Sellin, D.~Paleressompoulle, F.~Massi, M.~Simoneau, L.~Mouchnino, and J.~Blouin, ``Tactile contribution extends beyond exteroception during spatially guided finger movements,'' \emph{Scientific Reports}, vol.~15, no.~1, p. 14959, 2025.

\bibitem{frumento2024unconscious}
S.~Frumento, G.~Preatoni, L.~Chee, A.~Gemignani, F.~Ciotti, D.~Menicucci, and S.~Raspopovic, ``Unconscious multisensory integration: behavioral and neural evidence from subliminal stimuli,'' \emph{Frontiers in Psychology}, vol.~15, p. 1396946, 2024.

\bibitem{pandey2017mobile}
A.~Pandey, S.~Pandey, and D.~Parhi, ``Mobile robot navigation and obstacle avoidance techniques: A review,'' \emph{Int Rob Auto J}, vol.~2, no.~3, p. 00022, 2017.

\bibitem{lee2020making}
M.~A. Lee, Y.~Zhu, P.~Zachares, M.~Tan, K.~Srinivasan, S.~Savarese, L.~Fei-Fei, A.~Garg, and J.~Bohg, ``Making sense of vision and touch: Learning multimodal representations for contact-rich tasks,'' \emph{IEEE Transactions on Robotics}, vol.~36, no.~3, pp. 582--596, 2020.

\bibitem{guzey2023dexterity}
I.~Guzey, B.~Evans, S.~Chintala, and L.~Pinto, ``Dexterity from touch: Self-supervised pre-training of tactile representations with robotic play,'' \emph{arXiv preprint arXiv:2303.12076}, 2023.

\bibitem{ablett2024multimodal}
T.~Ablett, O.~Limoyo, A.~Sigal, A.~Jilani, J.~Kelly, K.~Siddiqi, F.~Hogan, and G.~Dudek, ``Multimodal and force-matched imitation learning with a see-through visuotactile sensor,'' \emph{IEEE T-RO}, 2024.

\bibitem{zhang2025kinedex}
D.~Zhang, C.~Yuan, C.~Wen, H.~Zhang, J.~Zhao, and Y.~Gao, ``Kinedex: Learning tactile-informed visuomotor policies via kinesthetic teaching for dexterous manipulation,'' \emph{arXiv preprint arXiv:2505.01974}, 2025.

\bibitem{chen2025dexforce}
C.~Chen, Z.~Yu, H.~Choi, M.~Cutkosky, and J.~Bohg, ``Dexforce: Extracting force-informed actions from kinesthetic demonstrations for dexterous manipulation,'' \emph{arXiv preprint arXiv:2501.10356}, 2025.

\bibitem{liu2024forcemimic}
W.~Liu, J.~Wang, Y.~Wang, W.~Wang, and C.~Lu, ``Forcemimic: Force-centric imitation learning with force-motion capture system for contact-rich manipulation,'' \emph{arXiv preprint arXiv:2410.07554}, 2024.

\bibitem{liu2025vitamin}
F.~Liu, C.~Li, Y.~Qin, A.~Shaw, J.~Xu, P.~Abbeel, and R.~Chen, ``Vitamin: Learning contact-rich tasks through robot-free visuo-tactile manipulation interface,'' \emph{arXiv preprint arXiv:2504.06156}, 2025.

\bibitem{zhao2025polytouch}
J.~Zhao, N.~Kuppuswamy, S.~Feng, B.~Burchfiel, and E.~Adelson, ``Polytouch: A robust multi-modal tactile sensor for contact-rich manipulation using tactile-diffusion policies,'' \emph{arXiv preprint arXiv:2504.19341}, 2025.

\bibitem{lin2024learning}
T.~Lin, Y.~Zhang, Q.~Li, H.~Qi, B.~Yi, S.~Levine, and J.~Malik, ``Learning visuotactile skills with two multifingered hands,'' \emph{arXiv preprint arXiv:2404.16823}, 2024.

\bibitem{huang2021visual}
B.~Huang, S.~D. Han, J.~Yu, and A.~Boularias, ``Visual foresight trees for object retrieval from clutter with nonprehensile rearrangement,'' \emph{IEEE RA-L}, vol.~7, no.~1, pp. 231--238, 2021.

\bibitem{yuan2019end}
W.~Yuan, K.~Hang, D.~Kragic, M.~Y. Wang, and J.~A. Stork, ``End-to-end nonprehensile rearrangement with deep reinforcement learning and simulation-to-reality transfer,'' \emph{Robotics and Autonomous Systems}, vol. 119, pp. 119--134, 2019.

\bibitem{muhayyuddin2017randomized}
Muhayyuddin, M.~Moll, L.~Kavraki, J.~Rosell, \emph{et~al.}, ``Randomized physics-based motion planning for grasping in cluttered and uncertain environments,'' \emph{IEEE RA-L}, vol.~3, no.~2, pp. 712--719, 2017.

\bibitem{srivastava2014combined}
S.~Srivastava, E.~Fang, L.~Riano, R.~Chitnis, S.~Russell, and P.~Abbeel, ``Combined task and motion planning through an extensible planner-independent interface layer,'' in \emph{IEEE ICRA}, 2014, pp. 639--646.

\bibitem{kurenkov2020visuomotor}
A.~Kurenkov, J.~Taglic, R.~Kulkarni, M.~Dominguez-Kuhne, A.~Garg, R.~Mart{\'\i}n-Mart{\'\i}n, and S.~Savarese, ``Visuomotor mechanical search: Learning to retrieve target objects in clutter,'' in \emph{IEEE IROS}.\hskip 1em plus 0.5em minus 0.4em\relax IEEE, 2020, pp. 8408--8414.

\bibitem{agboh2018real}
W.~C. Agboh and M.~R. Dogar, ``Real-time online re-planning for grasping under clutter and uncertainty,'' in \emph{IEEE Humanoids}.\hskip 1em plus 0.5em minus 0.4em\relax IEEE, 2018, pp. 1--8.

\bibitem{li2022see}
H.~Li, Y.~Zhang, J.~Zhu, S.~Wang, M.~A. Lee, H.~Xu, E.~Adelson, L.~Fei-Fei, R.~Gao, and J.~Wu, ``See, hear, and feel: Smart sensory fusion for robotic manipulation,'' \emph{arXiv preprint:2212.03858}, 2022.

\bibitem{murali2022active}
P.~K. Murali, A.~Dutta, M.~Gentner, E.~Burdet, R.~Dahiya, and M.~Kaboli, ``Active visuo-tactile interactive robotic perception for accurate object pose estimation in dense clutter,'' \emph{IEEE Robotics and Automation Letters}, vol.~7, no.~2, pp. 4686--4693, 2022.

\bibitem{gupta2013interactive}
M.~Gupta, T.~R{\"u}hr, M.~Beetz, and G.~S. Sukhatme, ``Interactive environment exploration in clutter,'' in \emph{IEEE IROS}, 2013, pp. 5265--5272.

\bibitem{huang2021mechanical}
H.~Huang, M.~Dominguez-Kuhne, V.~Satish, M.~Danielczuk, K.~Sanders, J.~Ichnowski, A.~Lee, A.~Angelova, V.~Vanhoucke, and K.~Goldberg, ``Mechanical search on shelves using lateral access x-ray,'' in \emph{IROS}.\hskip 1em plus 0.5em minus 0.4em\relax IEEE, 2021, pp. 2045--2052.

\bibitem{huang2022mechanical}
H.~Huang, , M.~Danielczuk, C.~M. Kim, L.~Fu, Z.~Tam, J.~Ichnowski, A.~Angelova, B.~Ichter, and K.~Goldberg, ``Mechanical search on shelves using a novel “bluction” tool,'' in \emph{IEEE ICRA}, 2022, pp. 6158--6164.

\bibitem{liu2025fetchbot}
W.~Liu, Y.~Wan, J.~Wang, Y.~Kuang, X.~Shi, H.~Li, D.~Zhao, Z.~Zhang, and H.~Wang, ``Fetchbot: Object fetching in cluttered shelves via zero-shot sim2real,'' \emph{arXiv preprint arXiv:2502.17894}, 2025.

\bibitem{bejjani2021occlusion}
W.~Bejjani, W.~C. Agboh, M.~R. Dogar, and M.~Leonetti, ``Occlusion-aware search for object retrieval in clutter,'' in \emph{2021 IEEE/RSJ International Conference on Intelligent Robots and Systems (IROS)}.\hskip 1em plus 0.5em minus 0.4em\relax IEEE, 2021, pp. 4678--4685.

\bibitem{thomasson2022going}
R.~Thomasson, E.~Roberge, M.~R. Cutkosky, and J.-P. Roberge, ``Going in blind: Object motion classification using distributed tactile sensing for safe reaching in clutter,'' in \emph{IEEE IROS}, 2022, pp. 1440--1446.

\bibitem{albini2021exploiting}
A.~Albini, F.~Grella, P.~Maiolino, and G.~Cannata, ``Exploiting distributed tactile sensors to drive a robot arm through obstacles,'' \emph{IEEE RA-L}, vol.~6, no.~3, pp. 4361--4368, 2021.

\bibitem{bhattacharjee2014robotic}
T.~Bhattacharjee, P.~M. Grice, A.~Kapusta, M.~D. Killpack, D.~Park, and C.~C. Kemp, ``A robotic system for reaching in dense clutter that integrates model predictive control, learning, haptic mapping, and planning.''\hskip 1em plus 0.5em minus 0.4em\relax Georgia Institute of Technology, 2014.

\bibitem{jain2013reaching}
A.~Jain, M.~D. Killpack, A.~Edsinger, and C.~C. Kemp, ``Reaching in clutter with whole-arm tactile sensing,'' \emph{IJRR}, vol.~32, no.~4, pp. 458--482, 2013.

\bibitem{bohg2010strategies}
J.~Bohg, M.~Johnson-Roberson, M.~Björkman, and D.~Kragic, ``Strategies for multi-modal scene exploration,'' in \emph{2010 IEEE/RSJ International Conference on Intelligent Robots and Systems}, 2010, pp. 4509--4515.

\bibitem{brouwer2024tactile}
D.~Brouwer, J.~Citron, H.~Choi, M.~Lepert, M.~Lin, J.~Bohg, and M.~Cutkosky, ``Tactile-informed action primitives mitigate jamming in dense clutter,'' \emph{IEEE ICRA (Accepted)}, 2024.

\bibitem{zhong2022soft}
S.~Zhong, N.~Fazeli, and D.~Berenson, ``Soft tracking using contacts for cluttered objects to perform blind object retrieval,'' \emph{IEEE RA-L}, vol.~7, no.~2, pp. 3507--3514, 2022.

\bibitem{pai2023tactofind}
S.~Pai, T.~Chen, M.~Tippur, E.~Adelson, A.~Gupta, and P.~Agrawal, ``Tactofind: A tactile only system for object retrieval,'' \emph{arXiv preprint arXiv:2303.13482}, 2023.

\bibitem{chi2023diffusion}
C.~Chi, Z.~Xu, S.~Feng, E.~Cousineau, Y.~Du, B.~Burchfiel, R.~Tedrake, and S.~Song, ``Diffusion policy: Visuomotor policy learning via action diffusion,'' \emph{IJRR}, p. 02783649241273668, 2023.

\bibitem{choi2022deep}
H.~Choi, D.~Brouwer, M.~A. Lin, K.~T. Yoshida, C.~Rognon, B.~Stephens-Fripp, A.~M. Okamura, and M.~R. Cutkosky, ``Deep learning classification of touch gestures using distributed normal and shear force,'' in \emph{IEEE IROS}, 2022.

\bibitem{huh2021multi}
T.~M. Huh, K.~Sanders, M.~Danielczuk, M.~Li, Y.~Chen, K.~Goldberg, and H.~S. Stuart, ``A multi-chamber smart suction cup for adaptive gripping and haptic exploration,'' in \emph{2021 IEEE IROS}.\hskip 1em plus 0.5em minus 0.4em\relax IEEE, 2021, pp. 1786--1793.

\bibitem{gu2025tactilealoha}
N.~Gu, K.~Kosuge, and M.~Hayashibe, ``Tactilealoha: Learning bimanual manipulation with tactile sensing,'' \emph{IEEE RA-L}, 2025.

\bibitem{de2019causal}
P.~De~Haan, D.~Jayaraman, and S.~Levine, ``Causal confusion in imitation learning,'' \emph{NeurIPS}, vol.~32, 2019.

\bibitem{hao2023masked}
Y.~Hao, R.~Wang, Z.~Cao, Z.~Wang, Y.~Cui, and D.~Sadigh, ``Masked imitation learning: Discovering environment-invariant modalities in multimodal demonstrations,'' in \emph{IEEE IROS}.\hskip 1em plus 0.5em minus 0.4em\relax IEEE, 2023, pp. 1--7.

\bibitem{kumar2025collage}
S.~Kumar, S.~Dass, G.~Pavlakos, and R.~Mart{\'\i}n-Mart{\'\i}n, ``Collage: Adaptive fusion-based retrieval for augmented policy learning,'' \emph{arXiv preprint arXiv:2508.01131}, 2025.

\bibitem{du2024haptiknit}
C.~du~Pasquier, L.~Tessmer, I.~Scholl, L.~Tilton, T.~Chen, S.~Tibbits, and A.~Okamura, ``Haptiknit: Distributed stiffness knitting for wearable haptics,'' \emph{Science Robotics}, vol.~9, no.~97, p. eado3887, 2024.

\bibitem{lu2024robot}
Z.~Lu, K.~Feng, J.~Xu, H.~Chen, and Y.~Lou, ``Robot safe planning in dynamic environments based on model predictive control using control barrier function,'' \emph{arXiv preprint arXiv:2404.05952}, 2024.

\bibitem{vinter2024safe}
F.~Vinter-Hviid, C.~Sloth, T.~R. Savarimuthu, and I.~Iturrate, ``Safe contact-based robot active search using bayesian optimization and control barrier functions,'' \emph{Frontiers in Robotics and AI}, vol.~11, p. 1344367, 2024.

\bibitem{haldar2023teach}
S.~Haldar, J.~Pari, A.~Rai, and L.~Pinto, ``Teach a robot to fish: Versatile imitation from one minute of demonstrations,'' \emph{arXiv preprint arXiv:2303.01497}, 2023.

\bibitem{xue2025reactive}
H.~Xue, J.~Ren, W.~Chen, G.~Zhang, Y.~Fang, G.~Gu, H.~Xu, and C.~Lu, ``Reactive diffusion policy: Slow-fast visual-tactile policy learning for contact-rich manipulation,'' \emph{arXiv preprint arXiv:2503.02881}, 2025.

\bibitem{killpack2016model}
M.~D. Killpack, A.~Kapusta, and C.~C. Kemp, ``Model predictive control for fast reaching in clutter,'' \emph{Autonomous Robots}, vol.~40, pp. 537--560, 2016.

\end{thebibliography}

\end{document}